\documentclass{article}

\usepackage[final]{neurips_2021}

\usepackage[utf8]{inputenc} %
\usepackage[T1]{fontenc}    %
\usepackage{url}            %
\usepackage{amsfonts}       %
\usepackage{nicefrac}       %
\usepackage{microtype}      %
\usepackage{algorithm}
\usepackage{algorithmic}
\usepackage{sidecap}
\usepackage[singlelinecheck=off]{caption}
\newlength\Myfigwd
\usepackage[export]{adjustbox}

\sidecaptionvpos{figure}{c}

\makeatletter
\newcommand\footnoteref[1]{\protected@xdef\@thefnmark{\ref{#1}}\@footnotemark}
\makeatother

\usepackage{amssymb}
\usepackage{amsthm}
\usepackage{url}
\usepackage{enumerate}
\usepackage{mathtools}
\usepackage{stmaryrd}
\usepackage{microtype}
\usepackage{graphicx}
\usepackage{subfigure}
\usepackage{color}
\usepackage{booktabs} %
\usepackage{subfigure}
\usepackage{multirow}
\usepackage{lipsum, babel}
\usepackage{verbatim}
\usepackage{xcolor, colortbl}
\usepackage{hyperref}
\usepackage{wrapfig}

\definecolor{Gray}{gray}{0.85}

\newcolumntype{a}{>{\columncolor{Gray}}c}

\newcommand{\X}{\mathcal{X}}

\newcommand{\Y}{\mathcal{Y}}
\newcommand{\R}{\mathbb{R}}
\newcommand{\spec}{\boldsymbol{\psi}}
\newcommand{\N}{\mathcal{N}}

\newcommand{\xnom}{x_{\text{nom}}}
\newcommand{\Support}[1]{\text{Support}\br{#1}}

\newcommand{\softmax}[1]{\texttt{softmax}\br{#1}}
\newcommand{\Pout}{\Pcal_{out}}

\newtheorem{theorem}{Theorem}

\newtheorem{proposition}{Proposition}
\newtheorem{definition}{Definition}

\newcommand{\br}[1]{\left({#1}\right)}
\newcommand{\norm}[1]{\left\|{#1}\right\|}

\newcommand{\relu}{\mathrm{\texttt{ReLU}}}
\DeclareMathOperator*{\EE}{\mathbb{E}}
\newcommand{\ExP}[2]{\EE_{#2}\left[{#1}\right]}

\newcommand{\diagmat}[1]{\mathrm{diag}\br{#1}}
\newcommand{\Sym}{\mathbb{S}}
\newcommand{\Pcal}{\mathcal{P}}
\newcommand{\pdrop}{p_{\text{dropout}}}

\newcommand{\One}{\mathbf{1}}

\newcommand{\xood}{x_{ood}}
\newcommand{\pw}{p^w}

\title{Make Sure You're Unsure: A Framework for Verifying Probabilistic Specifications}

\author{%
  Leonard Berrada\thanks{Equal contribution. Authors listed in alphabetical order. Correspondance to \texttt{lberrada@deepmind.com}, \texttt{sdathath@deepmind.com}, \texttt{dvij@cs.washington.edu}.} \thanks{DeepMind, London, United Kingdom.}
  \And
Sumanth Dathathri \footnotemark[1] \footnotemark[2] \And
Krishnamurthy (Dj) Dvijotham\footnotemark[1] \footnotemark[2] 
 \And
Robert Stanforth \footnotemark[2]  \And
Rudy Bunel \footnotemark[2] \And
Jonathan Uesato \footnotemark[2] \And
Sven Gowal \footnotemark[2] \And
M. Pawan Kumar \footnotemark[2]
}

\begin{document}

\maketitle

\begin{abstract}
Most real world applications require dealing with stochasticity like sensor noise or predictive uncertainty, where formal specifications of desired behavior are inherently probabilistic. 
Despite the promise of formal verification in ensuring the reliability of neural networks, progress in the direction of 
probabilistic specifications has been limited.
In this direction, we first introduce a general formulation of probabilistic specifications for neural networks, which captures both probabilistic networks (e.g., Bayesian neural networks, MC-Dropout networks) and uncertain inputs (distributions over inputs arising from sensor noise or other perturbations). 
We then propose a general technique to verify such specifications by generalizing the notion of Lagrangian duality, replacing standard Lagrangian multipliers with "functional multipliers" that can be arbitrary functions of the activations at a given layer. 
We show that an optimal choice of functional multipliers leads to exact verification (i.e.,  sound and complete verification),  and 
for specific forms of multipliers, we develop tractable practical verification algorithms.

We empirically validate our algorithms by applying them to Bayesian Neural Networks (BNNs) and MC Dropout Networks, and certifying properties such as adversarial robustness and robust detection of out-of-distribution (OOD) data.
On these tasks we are able to provide significantly stronger guarantees when compared to prior work -- for instance, for a VGG-64 MC-Dropout CNN trained on CIFAR-10 in a verification-agnostic manner, 
we improve the certified AUC (a verified lower bound on the true AUC) for robust OOD detection (on CIFAR-100) from $0 \% \rightarrow 29\%$.
Similarly, for a BNN trained on MNIST, we improve on the $\ell_\infty$ robust accuracy from 
$60.2 \% \rightarrow 74.6\%$.
Further, on a novel specification -- distributionally robust OOD detection -- we  
improve on the certified AUC from $5\% \rightarrow 23\%$.

\end{abstract}
\section{Introduction}

\label{intro}
While neural networks (NNs) have shown significant promise in a wide-range of applications (for e.g., 
\citep{resnet, autospeech}), a key-bottleneck towards their wide-spread adoption in 
safety-critical applications is the lack of formal guarantees regarding safety and performance.
In this direction, there has been considerable progress towards developing scalable
 methods that can provide formal guarantees regarding the conformance of NNs 
 with desired properties \citep{katz2017reluplex, dvijotham2018dual, raghunathan2018semidefinite}.
However, much of this progress has been in the setting where the specifications and 
neural networks do not exhibit any probabilistic behaviour, or is mostly
specialized for specific probabilistic specifications \citep{weng2019proven, wicker2020probabilistic}.
In contrast, we introduce a general framework for verifying specifications of neural networks 
that are probabilistic.
The framework enables us to handle stochastic neural networks such as Bayesian Neural 
Networks or Monte-Carlo (MC) dropout networks, as well as 
probabilistic properties, such as distributionally robust out-of-distribution (OOD) detection.
Furthermore, the specification can be defined on the output distribution from the 
network, which allows us to handle operations such as the expectation on functions of the neural network output.

Probabilistic specifications are relevant and natural to many practical problems. 
For instance, for robotics applications, there is uncertainty arising from noisy measurements from sensors, and 
uncertainty regarding the actions of uncontrolled agents (e.g. uncertainty regarding the behaviour of pedestrians for a self-driving vehicle).
Often these uncertainties are modelled using a probabilistic approach, where a distribution is specified (or possibly learnt) 
over the feasible set of events \citep{probabilisticrobotics}.
In such cases, we want to  provide guarantees regarding the network's conformance to desired properties 
in the distributional setting (e.g. given a model of the pedestrian's uncertain behaviour, 
guarantee that the probability of collision for the autonomous vehicle is small). 
A more general problem includes scenarios where there is uncertainty regarding the parameters of the 
distribution used to model uncertainty.
Here, in this general setting, we seek to verify the property that the network behaviour conforms with the desired specification 
under uncertainty corresponding to an entire set of distributions.

The key to handling the aforementioned complexity in the specifications being verified through our framework is the generalization of the Lagrangian duality. 
Specifically, instead of using the standard Lagrange duality where the multipliers are linear, we allow for probabilistic constraints (constraints between distributions) and use functional multipliers to replace the linear Lagrange multipliers.
This allows us to exploit the structure of these probabilistic constraints, enabling us to provide stronger guarantees and 
facilitates the verification of \emph{verification-agnostic networks} (networks that are not designed to be verifiable). 
In our paper, we focus on verification-agnostic networks as this is desirable for many reasons, as noted in 
\cite{dathathri2020enabling}.
To summarize, 
our main contributions are: 

\begin{itemize}
  \setlength\itemsep{0.0em}
  
    \item We derive a general framework that extends Lagrangian duality to handle a wide range of probabilistic specifications. Our main theoretical result (Theorem 1) shows that our approach (i) is always sound and computes an upper bound on the maximum violation of the specification being verified, and (ii) is expressive enough to theoretically capture tight verification (i.e. obtaining both sound and complete verification).
  
    \item We develop novel algorithms for handling specific multipliers and objectives within our framework (Propositions \ref{prop:linexp}, \ref{thm:softmax}). This allows us to apply our framework to novel specifications (such as distributionally robust OOD detection, where input perturbations are drawn from entire 
    sets of distributions) by better capturing the probabilistic structure of the problem.
    \item We empirically validate our method by verifying neural networks, which are verification-agnostic, on a variety of probabilistic specifications. We demonstrate that even with relatively simple choices for the functional multiplier, our method strongly outperforms prior methods, which sometimes provide vacuous guarantees only.
    This further points towards the potential for significant improvements to be had by developing tractable optimization techniques for more complex and expressive multipliers within our framework.
\end{itemize}
\section{Probabilistic Specifications}\label{sec:specs}
\subsection{Notation}

Let us consider a possibly stochastic neural network $\phi: \X \to \Pcal\br{\Y}$, where $\X$ is the set of possible input values to the model, $\Y$ is the set of possible output values, and $\Pcal\br{\Y}$ is the set of distributions over $\Y$. 
We assume that $\Y$ is a subset of $\R^l$ (unless specified otherwise), where $l$ is the number of labels, and the output of the model are logits corresponding to unnormalized log-confidence scores assigned to the labels $\{1, \ldots, l\}$.

The model is assumed to be a sequence of $K$ layers, each of them possibly stochastic.
For $k \in \{1, \ldots, K\}$, $\pi_k\br{x_{k}|x_{k-1}}$ denotes the probability that the output of layer $k$ takes value $x_k$ when its input value is $x_{k-1}$.
We write $x_{k} \sim \pi_k\br{x_{k-1}}$ to denote that $x_k$ is drawn from the distribution over outputs of layer $k$ given input $x_{k-1}$ to layer $k$.
We further assume that each $\pi_k\br{x}$ has the form $\sigma(\tilde{w}x+\tilde{b})$, where $\sigma$ is a non-linear activation function (e.g., $\relu$, sigmoid, $\texttt{MaxOut}$), and $\tilde{w}$ and $\tilde{b}$ are random variables. 
The stochasticity for layer $\pi_k$  is assumed to be statistically independent of the stochasticity at other layers.
For a BNN, $\tilde{w}$ and $\tilde{b}$ follow a diagonal Gaussian distribution (i.e., a Gaussian distribution with a diagonal covariance matrix), and for a MC-Dropout network they follow a Bernoulli-like distribution. 

Given a distribution $p_0$ over the inputs $\X$, we use $\phi(p_0)$ to denote (with a slight abuse) the distribution of the random variable $\phi(X_0)$, where $X_0 \sim p_0$.

\subsection{Problem Formulation.}
We now introduce the general problem formulation for which we develop the 
verification framework.
\begin{definition}[Probabilistic verification problem]
Given a (possibly stochastic) neural network $\phi: \X \to \Pcal\br{\Y}$, a set of distributions over the input $\mathcal{P}_0$ and a functional $\spec: \mathcal{P}\br{\Y} \mapsto \R$, the probabilistic verification problem is to check that the following is true:
\begin{align}
\forall \: p_0 \in \mathcal{P}_0, \: \spec\br{\phi\br{p_0}} \leq 0.\label{eq:spec}
\end{align}
\end{definition}
 
\subsection{Examples of Specifications}\label{sec:example_specifications}
Below we provide examples of probabilistic specifications 
which are captured by the above problem formulation, and that 
we further empirically validate our 
framework on.
In Appendix \ref{supp:specform}, 
we provide further examples of relevant specifications (e.g., ensuring reliable uncertainty calibration) that can be handled by our problem setup.
\label{sec:probabexamples}
\paragraph{Distributionally Robust OOD Detection.} 
We consider the problem of verifying that a stochastic neural network assigns low confidence scores to all labels for OOD inputs, even in the presence of bounded noise perturbations to the inputs.
Given a noise distribution perturbing an OOD image $x_\mathrm{ood}$,
we require that the expected softmax is smaller than a specified confidence threshold $p_{\max}$ for each label $i$.
Since the precise noise distribution is most often unknown, we wish to consider an entire class $\mathcal{P}_{noise}$ of 
noise distributions. 
Denoting by $\delta_x$ the Dirac distribution around $x$, the problem is then to guarantee that for every $p_0$ in \mbox{$\mathcal{P}_0 = \{\delta_{\xood} + \omega: \omega \in \mathcal{P}_{noise}\}$} and for each possible label $i$, $\spec(\phi(p_0)) \coloneqq \mathbb{E}_{y \sim \phi(p_0)}[\softmax{y}_i] - p_{\max} \leq 0$.
Robust OOD detection under bounded $\ell_\infty$ perturbations as considered in \cite{bitterwolf2020provable}
is a special case of this problem where $\mathcal{P}_{noise}$ is restricted to a set of $\delta$ distributions over points with bounded $\ell_\infty$ norm.

\paragraph{Robust Classification.} 
We also extend the commonly studied robust classification problem \citep{madry2017towards} under norm-bounded  perturbations, to the setting of probabilistic neural networks (e.g. BNNs). 
Define $\mathcal{P}_0$ to be the set of $\delta$ input distributions centered at points within an $\epsilon$-ball of a nominal point $\xnom$, with label $i \in \{1, \ldots, l\}$: $\mathcal{P}_0 = \{\delta_x: \norm{x-\xnom} \leq \epsilon\}$.
For every $p_0 \in \mathcal{P}_0$, we wish to guarantee that the 
stochastic NN correctly classifies the input, i.e. for each $j$, 
 $\spec(\phi(p_0)) \coloneqq \mathbb{E}_{y \sim \phi(p_0)}[\softmax{y}_i - \softmax{y}_j] \leq 0$.
Note that it is important to take the expectation of the softmax (and not logits) since this is how inference from BNNs is performed.

\section{The Functional Lagrangian Framework}
\label{sec:framework}
We consider the following optimization version:
\begin{align}
    \texttt{OPT} = \max_{p_0 \in \mathcal{P}_0} \spec\br{\phi\br{p_0}}, \label{eq:specopt}
\end{align}
Having $\texttt{OPT} \leq 0$  here is equivalent to satisfying specification~(\ref{eq:spec}) .
However, solving problem~\eqref{eq:specopt} directly to global optimality is intractable in general, because it can possibly be a challenging nonlinear and stochastic optimization problem. 
However, to only verify that the specification is satisfied, it may suffice to compute an upper bound on $\texttt{OPT}$. 
Here, we describe how the functional Lagrangian framework allows to derive such bounds by decomposing the overall problem into smaller, easier sub-problems.
 
\subsection{General Framework}\label{sec:General_Functional}

Let $\X_k$ denote the feasible space of activations at layer $k$, and let $p_k$ denote the distribution of activations at layer $k$ when the inputs follow distribution $p_0$ (so that $p_K=\phi\br{p_0}$).

\paragraph{Assumptions.} In order to derive our verification framework, we make the following assumptions:\\
(A1): $\exists \: l_0 \leq u_0 \in \R^n$ such that for each input distribution $p_0 \in \Pcal_0$, $\Support{p_0} \subseteq \X_0=[l_0, u_0]$. \\
(A2): Each layer is such that if $x \in \X_k=[l_k, u_k]$, then $\Support{\pi_k\br{x}} \subseteq \X_{k+1}=[l_{k+1}, u_{k+1}] $.

Assumption (A1) is natural since the inputs to neural networks are bounded. Assumption (A2) can be restrictive in some cases: it requires that the layer output is bounded with probability 1, which is not true, for example, if we have a BNN with a Gaussian posterior. However, we can relax this assumption to requiring that the output is bounded with high probability, as in \citet{wicker2020probabilistic}.

\paragraph{Functional Lagrangian Dual.}
In order to derive the dual, we begin by noting that problem \eqref{eq:specopt} can be equivalently written in the following constrained form:
\begin{equation*} \label{eq:pb_constrained_form}
    \max_{p_0 \in \mathcal{P}_0, p_1, \ldots, p_K} \: \spec\br{p_K} \text{s.t. } \forall \: k \in \{0, \ldots, K-1\}, \forall \: y \in \X_{k+1}, \: p_{k+1}\br{y} = \int_{\X_k} \pi_k\br{y|x} p_k\br{x}dx.
\end{equation*}
For the $k$-th constraint, let us assign a Lagrangian multiplier $\lambda_{k+1}(y)$ to each possible $y \in \X_{k+1}$. 
Note that $\lambda\br{y}$ is chosen independently for each $y$, hence $\lambda$ is a \emph{functional multiplier}. 
We then integrate over $y$, which yields the following Lagrangian penalty to be added to the dual objective:
\begin{equation}
     -\int_{\X_{k+1}} \lambda_{k+1}\br{y}p_{k+1}\br{y}dy  + \displaystyle\int_{\X_k, \X_{k+1}} \lambda_{k+1}\br{y}\pi_k\br{y|x}p_k\br{x}dx dy.
\end{equation}
We now make two observations, which are described here at a high level only and are available in more details in appendix \ref{app:funlag_thm}.
First, if we sum these penalties over $k$ and group terms by $p_k$, it can be observed that the objective function decomposes additively over the $p_k$ distributions.
Second, for $k \in \{1, \ldots, K-1\}$, each $p_k$ can be optimized independently (since the objective is separable), and since the objective is linear in $p_k$ , the optimal $p_k$ is a Dirac distribution, which means that the search over the probability distribution $p_k$ can be simplified to a search over feasible values $x_k \in \X_k$.
This yields the following dual:
\begin{align}  \label{eq:dual_explicit}
    &\max_{p_K \in \mathcal{P}_K} \br{\spec\br{p_K}-\int_{\X_K} \lambda_K\br{x}p_K\br{x}dx} + \sum_{k=1}^{K-1} \max_{x \in \X_k} \bigg(\int_{\X_{k+1}} \lambda_{k+1}\br{y}\pi_k\br{y|x}dy - \lambda_{k}\br{x}\bigg) \notag \\
    &\quad  + \max_{p_0 \in \mathcal{P}_0} \int_{\X_0}\br{\int_{\X_1} \lambda_1\br{y}\pi_0\br{y|x}dy}p_0\br{x}dx,
\end{align}
where we define $\mathcal{P}_K \triangleq \phi(\mathcal{P}_0)$.
In the rest of this work, we refer to this dual as $g\br{\lambda}$, and we use the following notation to simplify equation \eqref{eq:dual_explicit}:
\begin{equation} 
g\br{\lambda} 
    = \max_{p_0 \in \mathcal{P}_0} g_0(p_0, \lambda_1) + \sum_{k=1}^{K-1} \max_{x_k \in \X_k} g_k(x_k, \lambda_k, \lambda_{k+1}) + \max_{p_K \in \mathcal{P}_K}  g_K(p_K, \lambda_K). \label{eq:thmdual}
\end{equation}
The dual $g\br{\lambda}$ can be seen as a generalization of Lagrangian relaxation \citep{bertsekas2015convex} with the two key modifications: (i) layer outputs are integrated over possible values, and (ii) Lagrangian penalties are expressed as arbitrary functions $\lambda_k\br{x}$ instead of being restricted to linear functions.

\paragraph{Main Result.} Here, we relate the functional Lagrangian dual to the specification objective~\eqref{eq:specopt}.

\begin{theorem}\label{thm:dual}
For any collection of functions $\lambda = (\lambda_1, \ldots, \lambda_K) \in \mathbb{R}^{\X_1} \times \ldots \times \mathbb{R}^{\X_K}$, we have that $g\br{\lambda} \geq \texttt{OPT}$.
In particular, if a choice of $\lambda$ can be found such that $g\br{\lambda} \leq 0$, then specification (\ref{eq:spec}) is true. 
Further, when $\spec\br{p_K} = \ExP{c\br{y}}{y \sim p_K}$, the dual becomes tight: $g\br{\lambda^\star} = \texttt{OPT}$ if $\lambda^\star$ is set to:
\begin{equation*}
\lambda^\star_K\br{x} = c\br{x};
\forall \: k \in \{K-1, \ldots, 1\}, \: \lambda^\star_k\br{x} = \ExP{\lambda^\star_{k+1}\br{y}}{y \sim \pi_k\br{x}}.
\end{equation*}
\end{theorem}
\begin{proof}
We give a brief sketch of the proof - the details are in Appendix \ref{app:funlag_thm}.
The problem in constrained form is an infinite dimensional optimization with decision variables $p_0, p_1, \ldots, p_K$ and linear constraints relating $p_k$ and $p_{k+1}$. 
The Lagrangian dual of this optimization problem has objective $g\br{\lambda}$.
By weak duality, we have $g(\lambda)\geq\texttt{OPT}$. 
The second part of the theorem is easily observed by plugging in $\lambda^\star$ in $g\br{\lambda}$ and observing that the resulting optimization problem is equivalent to \eqref{eq:specopt}.
\end{proof}

\paragraph{Example.}
Let $\mathcal{P}_0$ be the set of probability distributions with mean $0$, variance $1$, and support $[-1, 1]$, and let $\mathcal{N}_{[a, b]}(\mu, \sigma^2)$ denote the normal distribution with mean $\mu$ and variance $\sigma^2$ with truncated support $[a, b]$.
Now consider the following problem, for which we want to compute an upper bound:
\begin{equation}
 \texttt{OPT} = \max_{p_0 \in \mathcal{P}_0} \mathbb{E}_{X_1}[\exp(-X_1)]  \quad \text{s.t. } X_1|X_0 \sim \mathcal{N}_{[0, 1]}(X_0^2, 1) \text{ and } X_0 \sim p_0.
\end{equation}
This problem has two difficulties that prevent us from applying traditional optimization approaches like Lagrangian duality \citep{bertsekas2015convex}, which has been used in neural network verification \cite{dvijotham2018dual}.
The first difficulty is that the constraint linking $X_1$ to $X_0$ is stochastic, and standard approaches can not readily handle that.
Second, the optimization variable $p_0$ can take any value in an entire set of probability distributions, while usual methods can only search over sets of real values.
Thus standard methods fail to provide the tools to solve such a problem.
Since the probability distributions have bounded support, a possible way around this problem is to ignore the stochasticity of the problem, and to optimize over the worst-case realization of the random variable $X_1$ in order to obtain a valid upper bound on $\texttt{OPT}$ as:
$
\texttt{OPT}
\leq \max_{x_1 \in [0, 1]} \exp(-x_1)  = 1.
$
However this is an over-pessimistic modeling of the problem and the resulting upper bound is loose.
In contrast, Theorem \ref{thm:dual} shows that for any function $\lambda: \mathbb{R} \to \mathbb{R}$, $\texttt{OPT}$ can be upper bounded by:
\begin{align*}
\texttt{OPT}
&\leq \max_{x_1 \in [0, 1], p_0 \in \mathcal{P}_0} \exp(-x_1) - \lambda(x_1)  + \mathbb{E}_{X_0 \sim p_0} [\mathbb{E}_{X_1|X_0 \sim \mathcal{N}_{[0, 1]}(X_0^2, 1)}[\lambda(X_1)]].
\end{align*}
This inequality holds true in particular for any function $\lambda$ of the form $x \mapsto \theta x$ where $\theta \in \mathbb{R}$, and thus:
\begin{align*}
\texttt{OPT}
&\leq \inf_{\theta \in \mathbb{R}} \max_{x_1 \in [0, 1], p_0 \in \mathcal{P}_0} \exp(-x_1) - \theta x_1  + \mathbb{E}_{X_0 \sim p_0} [\mathbb{E}_{X_1|X_0 \sim \mathcal{N}_{[0, 1]}(X_0^2, 1)}[\theta X_1 ]], \\
&= \inf_{\theta \in \mathbb{R}} \max_{x_1 \in [0, 1], p_0 \in \mathcal{P}_0} \exp(-x_1) - \theta x_1  + \theta \mathbb{E}_{X_0 \sim p_0} [X_0^2], \\
&= \inf_{\theta \in \mathbb{R}} \max_{x_1 \in [0, 1]} \exp(-x_1) - \theta x_1 + \theta  \approx 0.37.
\end{align*}
Here, our framework lets us tractably compute a bound on $\texttt{OPT}$ that is significantly tighter compared to the naive support-based bound.

\subsection{Optimization Algorithm}
\label{sec:verify_practical}

\paragraph{Parameterization.} 
The choice of functional multipliers affects the difficulty of evaluating $g\br{\lambda}$. In fact, since neural network verification is NP-hard \citep{katz2017reluplex}, we know that computing $g\br{\lambda^\star}$ is intractable in the general case.
Therefore in practice, we instantiate the functional Lagrangian framework for specific parameterized classes of Lagrangian functions,  which we denote as $\lambda\br{\theta}=\left\{\lambda_k\br{x} = \lambda_k\br{x;\theta_k}\right\}_{k=1}^K$.
Choosing the right class of functions $\lambda(\theta)$ is a trade-off: for very simple classes (such as linear functions), $g(\lambda(\theta))$ is easy to compute but may be a loose upper bound on (\ref{eq:specopt}), while more expressive choices lead to tighter relaxation of (\ref{eq:specopt}) at the cost of more difficult evaluation (or bounding) of $g(\lambda(\theta))$.

\begin{algorithm}[ht]
   \caption{Verification with Functional Lagrangians}
   \label{alg:summary}
\begin{algorithmic}
   \STATE {\bfseries Input:} initial dual parameters $\theta^{(0)}$, learning-rate $\eta$, number of iterations $T$.
   \FOR[\texttt{\small optimization loop}]{$t=0, \ldots, T-1$}
   \FOR[\texttt{\small potentially in parallel}]{$k=0$ {\bfseries to} $K$}
   \STATE $d_\theta^{(k)} = \nabla_{\theta} \left[ \displaystyle{\max_{x_k}} g_k(x_k, \lambda_k, \lambda_{k+1}) \right]$ \COMMENT{\texttt{\small potentially approximate maximization}}
   \ENDFOR
   \STATE $\theta^{(t+1)} = \theta^{(t)} - \eta \sum_{k=0}^K d_\theta^{(k)}$ \COMMENT{\texttt{\small or any gradient based optimization}}
   \ENDFOR
  \STATE {\bfseries Return:} Exact value or guaranteed upper bound on $g(\lambda(\theta^{(T)}))$ \COMMENT{\texttt{\small final evaluation}}
\end{algorithmic}
\end{algorithm}

\paragraph{Optimization.} 
With some abuse of notation, for convenience, we write $g_0 \br{x_0, \lambda_0, \lambda_1} \coloneqq  g_0\br{p_0, \lambda_1} $ and $g_K\br{x_K, \lambda_K, \lambda_{K+1}} \coloneqq g_K\br{p_K, \lambda_K}$,
with $\lambda_0=\lambda_{K+1}=0$. 
Then the problem of obtaining the best bound can be written as: $\min_{\theta} \sum_{k=0}^K \max_{x_k} g_k\br{x_k, \lambda_k, \lambda_{k+1}}$, where the inner maximizations are understood to be performed over the appropriate domains ($\Pcal_0$  for $x_0$, $\X_k$ for $x_k$, $l=1, \ldots, K-1$ and $\Pcal_K$ for $x_K$).
The overall procedure is described in Algorithm \ref{alg:summary}: $\theta$ is minimized by a gradient-based method in the outer loop; in the inner loop, the decomposed maximization problems over the $x_k$ get solved, potentially in parallel.  
During optimization, the inner problems can be solved approximately as long as they provide sufficient information about the descent direction for~$\theta$. 

\paragraph{Guaranteeing the Final Results.}
For the final verification certificate to be valid, we do require the final evaluation to provide the exact value of $g(\lambda(\theta^{(T)}))$ or an upper bound.
In the following section, we provide an overview of novel bounds that we use in our experiments to certify the final results.

\subsection{Bounds for Specific Instantiations}\label{sec:inner_max}

The nature of the maximization problems encountered by the optimization algorithm depends on the verification problem as well as the type of chosen Lagrangian multipliers.
In some easy cases, like linear multipliers on a ReLU layer, this results in tractable optimization or even closed-form solutions.
In other cases however, obtaining a non-trivial upper bound is more challenging.
In this section, we detail two such situations for which novel results were required to get tractable bounds: distributionally robust verification and expected softmax-based problems.
To the best of our knowledge, these bounds do not appear in the literature and thus constitute a novel contribution.

\paragraph{Distributionally Robust Verification with Linexp Multipliers.} 
We consider the setting where we verify a deterministic network with stochastic inputs and constraints on the input distribution $p_0 \in \Pcal_0$. 
In particular, we consider $\Pcal_0 = \{\mu + \omega: \omega \sim \Pcal_{noise}\}$, where $\Pcal_{noise}$ denotes a class of zero-mean noise distributions that all satisfy the property of having sub-Gaussian tails (this is true for many common noise distributions including Bernoulli, Gaussian, truncated Gaussian):
\[ \text{Sub-Gaussian tail:} \quad \forall i, \forall t \in \R, \: \ExP{\exp\br{t\omega_i}}{} \leq \exp\br{t^2 \sigma^2 / 2}.\]
We also assume that each component of the noise $\omega_i$ is i.i.d. 
The functional Lagrangian dual $g\br{\lambda}$ only depends on the input distribution $p_0$ via $g_0$, which evaluates to $g_0\br{p_0, \lambda_1} = \ExP{\lambda_1\br{x}}{x \sim p_0}$.
If we choose $\lambda_1$ to be a linear or quadratic function, then $g\br{\lambda}$ only depends on the first and second moments of $p_0$. This implies that the verification results will be unnecessarily conservative as they don't use the full information about the distribution $p_0$. 
To consider the full distribution it suffices to add an exponential term which evaluates to the moment generating function of the input distribution. 
Therefore we choose $\lambda_1\br{x} = \alpha^T x + \exp\br{\gamma^T x+\kappa}$ and $\lambda_2\br{x}=\beta^T x$.
The following result then gives a tractable upper bound on the resulting maximization problems:
\begin{proposition}\label{prop:linexp}
In the setting described above, and with $s$ as the element-wise activation function:
\begin{align*}
& \max_{p_0 \in \Pcal_0} g_0\br{p_0, \lambda_1} \leq  \alpha^T\br{ w \mu + b} +  \exp\br{\norm{w^T\gamma}^2\sigma^2 / 2 + \gamma^T b + \kappa}, \\
&  \max_{x \in \X_1} g_1\br{x, \lambda_1, \lambda_2} \leq \max_{x \in \X_2, z = s\br{x}} \beta^T \br{w_2z+b_2} -\alpha^Tx -\exp\br{\gamma^T x + \kappa}.
\end{align*}
The maximization in the second equation can be bounded by solving a convex optimization problem (Appendix~\ref{app:LinExP}).
\end{proposition}

\paragraph{Expected Softmax Problems.}
Several of the specifications discussed in Section \ref{sec:example_specifications} (e.g., distributionally robust OOD detection) require us to bound the expected value of a linear function of the softmax.
For specifications whose function can be expressed as an expected value: $\spec\br{p_K}=\ExP{c\br{x}}{x \sim p_K}$, 
by linearity of the objective w.r.t. the output distribution $p_K$, the search over the distribution $p_k$ can be simplified to a search over feasible output values $x_K$: 
\begin{align}
\max_{p_K \in \mathcal{P}_K} \spec\br{p_K}-\int_{\X_K} \lambda_K\br{x}p_K\br{x}dx = \max_{x \in \X_K} c\br{x}-\lambda_K\br{x}. \label{eq:final_opt}    
\end{align}
Given this observation, the following lets us certify results for linear functions of the $\softmax{x}$:
\begin{proposition}
\label{thm:softmax}
For affine $\lambda_K$, and $c(x)$ with the following form \mbox{$c\br{x} = \mu^T \softmax{x}$},  
$\max_{x \in \X_K} c\br{x}-\lambda_K\br{x}$ can be computed in time $O(3^d)$, where $\X_K \subseteq \R^d$.
\end{proposition}
We provide a proof of this proposition and a concrete algorithm for computing the solution  in Appendix \ref{app:softmax_opt}. 
This setting is particularly important to measure verified confidence and thus to perform robust OOD detection. We further note that while the runtime is exponential in $d$, $d$ corresponds to the number of labels in classification tasks which is a constant value and does not grow with the size of the network or the inputs to the network. Further, the computation is embarassingly parallel and can be done in $O(1)$ time if $3^d$ computations can be run in parallel. 
For classification problems with $10$ classes (like CIFAR-10 and MNIST), exploiting this parallelism, we can solve these problems on the order of milliseconds on a cluster of CPUs.
\section{Related Work}

\label{sec:relatedwork}

\paragraph{Verification of Probabilistic Specifications.}
We recall that in our work, $\Pcal_0$ refers to a space of distributions on the inputs $x$ to a network $\phi$, and that we address the following problem: verify that $\forall p \in \Pcal_0, \: \phi\br{p} \in \Pout$, where $\Pout$ represents a constraint on the output distribution. %
In contrast, prior works by \citet{weng2019proven}, \citet{fazlyab2019probabilistic}, and \citet{mirman2020robustness} study probabilistic specifications that involve robustness to probabilistic perturbations of a single input for 
deterministic networks.
This setting can be recovered as a special case within our formalism by letting the class $\Pcal_0$ contain a single distribution $p$. 
Conversely, \citet{dvijotham2018verification} study specifications involving stochastic models, but can not handle stochasticity in the input space.

\citet{wicker2020probabilistic} define a notion of probabilistic safety for BNNs: \mbox{$
Prob_{w \sim \mathcal{P}_w} \left[\forall x \in \X , \: \phi_w(x) \in \mathcal{C}\right] \geq p_{\min}$},
where $\phi_w$ denotes the network with parameters $w$, $\mathcal{P}_w$ denotes the distribution over network weights (e.g., a Gaussian posterior) and $\mathcal{C}$ is a set of safe outputs, and this allows for computation of the probability that 
a randomly sampled set of weights exhibits safe behaviour.
However, in practice, inference for BNNs  is carried out by averaging over predictions under the distribution of network weights. 
In this less restrictive and more practical setting, it suffices if the constraint is satisfied by the probabilistic prediction that averages over sampled weights: $
 \forall x \in \X, \: Prob_{w \sim \mathcal{P}_w} \left[\phi_w(x) \in \mathcal{C} \right] \geq p_{\min}$,
where $\phi_w(x)$ denotes the distribution over outputs for $x \in \X$.
Further, \citet{wicker2020probabilistic} also observe that $\min_{x \in \X} Prob_{w \sim \mathcal{P}_w}  \left[\phi_w(x) \in \mathcal{C}\right] \geq Prob_{w \sim \mathcal{P}_w}  \left[\forall x \in \X, \: \phi_w(x) \in \mathcal{C} \right]$, making the second constraint less restrictive.
\citet{cardelli2019statistical} and \citet{michelmore2020uncertainty} consider a similar specification, but unlike the approaches used here and by \citet{wicker2020probabilistic}, these methods can give statistical confidence bounds but not certified guarantees.

\cite{wickercertadvbnn} improve the classification robustness of Bayesian neural networks by training them to be robust based on an empirical estimate of the average upper bound on the specification violation, for a fixed set of sampled weights. 
In contrast, our approach provides meaningful guarantees for BNNs trained without considerations to make them more easily verifiable, and the guarantees our framework provides hold for inference based on the true expectation, as opposed 
to a fixed set of sampled~weights.

Our work also generalizes \citet{bitterwolf2020provable}, which studies specifications of the output distribution of NNs when the inputs and network are deterministic. In contrast, our framework's flexibility allows for stochastic networks as well. Furthermore, while \citet{bitterwolf2020provable} are concerned with training networks to be verifiable, their verification method fails for networks trained in a verification-agnostic manner. In our experiments, we provide stronger guarantees for networks that are trained in a verification-agnostic manner. 

\paragraph{Lagrangian Duality.}
Our framework subsumes existing methods that employ Lagrangian duality for NN verification.
In Appendix \ref{app:dvijeq}, we show that the functional Lagrangian dual instantiated with linear multipliers is equivalent to the dual from \citet{dvijotham2018dual}. 
This is also the dual of the LP relaxation \citep{ehlers2017formal} and the basis for other efficient NN verification algorithms (\citep{zhang2018efficient, singh2018fast}, for example), as shown in \citet{liu2019algorithms}.
For the case of quadratic multipliers and a particular grouping of layers, we show that our framework is equivalent to the Lagrangian dual of the SDP formulation from \citet{raghunathan2018semidefinite} (see Appendix \ref{app:sdpcert}). 

We also note that similar mathematical ideas on nonlinear Lagrangians have been explored in the optimization literature \citep{nedich2008geometric, feizollahi2017exact} but only as a theoretical construct - this has not lead to practical algorithms that exploit the staged structure of optimization problems arising in NN verification. 
Further, these approaches do not handle stochasticity.

\section{Experiments}
\label{sec:experiments}

Here, we empirically validate the theoretical flexibility of the framework and its applicability to across several specifications and networks. 
Crucially, we show that our framework permits verification of probabilistic specifications by effectively handling parameter and input stochasticity across tasks.
For all experiments, we consider a layer decomposition of the network such that the intermediate inner problems can be solved in closed form with linear multipliers (See Appendix \ref{app:linearmults}).
We use different bound-propagation algorithms to compute activation bounds based on the task, and generally refer to these methods as BP.
Our code is available at \url{https://github.com/deepmind/jax_verify}.

\subsection{Robust OOD Detection on Stochastic Neural Networks}
\label{sec:robustood}
\begin{table}[t!]
  \centering
  \small
  \caption{
    Robust OOD Detection: MNIST vs EMNIST (MLP and LeNet) and CIFAR-10 vs CIFAR-100 (VGG-*).
    BP: Bound-Propagation (baseline); FL: Functional Lagrangian (ours).
    The reported times correspond to the median of the 500 samples.
    }
 \begin{tabular}{l cc c c | c c | c c| c}
  \toprule
        \multirow{2}{*}{OOD Task} &\multirow{2}{*}{Model} & \multirow{2}{*}{\#neurons} & \multirow{2}{*}{\#params} & \multirow{2}{*}{$\epsilon$} & \multicolumn{2}{c|}{Time (s)} & \multicolumn{2}{c|}{GAUC (\%)} & \multirow{2}{*}{AAUC (\%)} \\
        \cmidrule(lr){6-7} \cmidrule(lr){8-9}
         &  &  &  &  & BP & FL &BP &FL &\\
        \midrule
        \multirow{3}{*}{(E)MNIST} 
        & \multirow{3}{*}{MLP} & \multirow{3}{*}{256} & \multirow{3}{*}{2k} & 0.01 &40.1 & +38.8 & 55.4 & {\bf 65.0} & {\color{blue} 86.9} \\
         & &  &  & 0.03 &40.1 & +37.4 & 38.5 & {\bf 53.1} & {\color{blue} 88.6} \\
         & &  & & 0.05 &38.4 & +36.2 & 18.9 & {\bf 36.1} & {\color{blue} 88.8} \\ \midrule
        \multirow{3}{*}{(E)MNIST} 
        &\multirow{3}{*}{LeNet}  & \multirow{3}{*}{0.3M} & \multirow{3}{*}{0.1M} & 0.01 & 53.2& +52.4 & 0.0 & {\bf 29.8} & {\color{blue} 71.6} \\
         & & & & 0.03 & 52.4& +51.1 & 0.0 & {\bf 14.1} & {\color{blue} 57.6} \\
         & & & & 0.05 & 55.4& +54.1 & 0.0 & {\bf 3.1} & {\color{blue} 44.0} \\ \midrule
        \multirow{3}{*}{CIFAR}
        &VGG-16 &  3.0M & 83k & 0.001 & 50.8 & +35.0& 0.0 & {\bf 25.6} & {\color{blue} 60.9} \\
        &VGG-32 &  5.9M & 0.2M & 0.001 & 82.3 & +40.9& 0.0 & {\bf 25.8} & {\color{blue} 64.7} \\
        &VGG-64 &  11.8M & 0.5M & 0.001 & 371.7 & +48.7 & 0.0 & {\bf 29.5} & {\color{blue} 67.4} \\
        \bottomrule
    \end{tabular}
    \label{tab:ood_auc_stochastic}
\end{table}
\paragraph{Verification Task.} 
We consider the task of robust OOD detection for stochastic neural networks under bounded $\ell_\infty$ inputs perturbation with radius $\epsilon$.
More specifically, we wish to use a threshold on the softmax value (maximized across labels) to classify whether a sample is OOD.
By using verified upper bounds on the softmax value achievable under $\epsilon$-perturbations, we can build a detector that classifies OOD images as such even under $\epsilon$ perturbations.
We use the Area Under the Curve (AUC) to measure the performance of the detector.
Guaranteed AUC (GAUC) is obtained with verified bounds on the softmax, and Adversarial AUC (AAUC) is based on the maximal softmax value found through an adversarial attack.
We consider two types of stochastic networks:
i) Bayesian neural networks (BNNs) whose posterior distribution is a truncated Gaussian distribution.
We re-use the MLP with two hidden layers of 128 units from \cite{wicker2020probabilistic} (denoted as MLP) and truncate their Gaussian posterior distribution to three standard deviations,
ii) we consider MC-Dropout convolutional neural networks, namely we use LeNet  (as in \citet{gal2016dropout}) and VGG-style models \citep{Simonyan15}.
\paragraph{Method.} 
We use linear Lagrangian multipliers, which gives closed-form solutions for all intermediate inner maximization problems (Appendix \ref{app:linearmults}). 
In addition, we leverage Proposition \ref{thm:softmax} to solve the final inner problem with the softmax specification objective.
Further experimental details, including optimization hyper-parameters, are available in Appendix \ref{app:experimental_details_ood}.
We compute activation bounds based on an extension of \citet{bunel2020efficient} to the bilinear case, referred to as BP in Table \ref{tab:ood_auc_stochastic}.
The corresponding bounds are obtained with probability 1, and if we were to relax these guarantees to only hold up to some probability lower than 1, we note that the method of \citep{wicker2020probabilistic} would offer tighter bounds.
\paragraph{Results.} 
The functional Lagrangian (FL) approach systematically outperforms the Bound-Propagation (BP) baseline.
We note that in particular, FL significantly outperforms BP on dropout CNNs, where BP is often unable to obtain any guarantee at all (See Table \ref{tab:ood_auc_stochastic}).
As the size of the VGG model increases, we can observe that the median runtime of BP increases significantly, while the additional overhead of using FL remains modest.

\subsection{Adversarial Robustness for Stochastic Neural Networks}
\label{sec:experiments_adv_bnn}
\paragraph{Verification Task.}
For this task, we re-use the BNNs trained on MNIST \citep{mnist2010}  
from \cite{wicker2020probabilistic} 
(with the Gaussian posterior truncated to three standard deviations bounded).
We use the same setting as \cite{wicker2020probabilistic} and study the classification 
robustness under $\ell_\infty$ perturbations  for 1-layered BNNs and 2-layered BNN at 
radii of $\epsilon=0.025$ and $\epsilon=0.001$ respectively.
We recall, as mentioned earlier in Section \ref{sec:relatedwork},  that the specification we study is different from that studied in \citep{wicker2020probabilistic}.
\paragraph{Method}
We use the same solving methodology as in Section \ref{sec:robustood}.
To compute bounds on the activations, we use the bilinear LBP method proposed in \citet{wicker2020probabilistic}.
\paragraph{Results.} 
Across settings (Table \ref{tab:supp_bnn_adv}) our approach  is able to significantly improve on the guarantees provided by the LBP baseline, while also noting that our method incurs an increased compute cost.
\begin{table}[!t]
  \centering
  \small
  \caption{
  Adversarial Robustness for different BNN architectures trained on MNIST from \cite{wicker2020probabilistic}. 
   The accuracy reported for FL and LBP is the \% of samples we can certify as robust with  probability 1.
   For each model, we run the experiment for the first 500 test-set samples. }
  \begin{tabular}{c c c| c c  | c  c }
  \toprule
   \#layers  & $\epsilon$ &  \#neurons  & LBP Acc. (\%) & FL Acc. (\%)  & LBP Time (s) & FL Time (s) \\
        \midrule
      &    &  128 & 67.0 & \textbf{77.2} & 16.7 & +518.3 \\
     1 & 0.025 & 256 & 66.2 & \textbf{76.4} & 16.1 & +522.8 \\
      &  & 512 & 60.2 & \textbf{74.6} & 16.0  & +522.4\\
     \hline 
     &  & 256 & 57.0 & \textbf{70.0} & 16.8 & +516.5\\
    2 &  0.001 & 512 & 79.6 & \textbf{87.4} & 17.0 & +517.3\\
     &  & 1024 & 39.4 & \textbf{42.4} & 17.1 & +514.1\\ 
    \bottomrule
    \end{tabular}
    \label{tab:supp_bnn_adv}
\end{table}

\subsection{Distributionally Robust OOD Detection}
\label{sec:distood}

\paragraph{Verification Task.} We bound the largest softmax probability across all labels for OOD inputs over noisy perturbations of the input where the noise is drawn from a family of distributions with only two constraints for each $p \in \Pcal_{noise}$:
$\omega \in [-\epsilon, \epsilon] \text{ with prob. } 1, 
\ExP{\exp\br{\omega t}}{\omega \sim p} \leq \exp\br{\sigma^2 t^2/2}$, 
for given constraints $\epsilon, \sigma > 0$. The first constraint corresponds to a restriction on the support of the noise distribution, and the second constraint requires that the noise distribution is sub-Gaussian with parameter $\sigma$. 
We note that no prior verification method, to the best of our knowledge, addresses this setting. Thus, as a baseline, we use methods that only use the support of the distribution and perform verification with respect to worst-case noise realizations within these bounds.

\paragraph{Method.} 
We use a 3-layer CNN trained on MNIST with the approach from \cite{hein2019relu} (details in Appendix \ref{app:experimental_details_dist_ood}), and 
 use functional multipliers of the form: $\lambda_k\br{x}=\theta_k^Tx$ for $k > 1$ and linear-exponential multipliers for the input layer: $\lambda_1\br{x} = \theta_1^T x + \exp\br{\gamma_1^Tx + \kappa_1}$ (method denoted by FL-LinExp).
As baselines, we consider a functional Lagrangian setting with linear multipliers that only uses information about the expectation of the noise distribution (method denoted by FL-Lin), and a BP baseline that only uses information about the bounds on the support of the noise distribution $(-\epsilon, \epsilon)$ (activation bounds are computed using \citet{bunel2020efficient}).
The inner maximization of $g_k$ for $K-1 \geq k \geq 2$ can be done in closed form and we use approaches described in Propositions \ref{prop:linexp} and \ref{thm:softmax} to respectively solve $\max g_0, \max g_1$ and $\max g_K$.
We use $\epsilon=0.04, \sigma=0.1$.
\begin{table}[ht!]
    \centering
    \small
    \caption{ Guaranteed Area Under Curve (GAUC) values in a distributionally robust setting. The stochastic formulation of the Functional Lagrangian with Linear-Exponential (LinExp) multipliers gets the highest guaranteed AUC. }\label{tab:label_oodauc_dist}
    \begin{tabular}{l c |c c c | c c c}
    \toprule 
    \multirow{2}{*}{Model} &\multirow{2}{*}{\#neurons} &\multicolumn{3}{c|}{GAUC (\%)} & \multicolumn{3}{c}{Timing (s)} \\
    \cmidrule(lr){3-5} \cmidrule(lr){6-8}
    && BP & FL-Lin &  FL-LinExp &BP & FL-Lin &  FL-LinExp \\
    \midrule
    CNN & 9972 &5.4 & 5.6 & \textbf{23.2} &227.7 &+661.8 &+760.7\\
    \bottomrule
    \end{tabular}
\end{table}

\paragraph{Results.}  
FL-LinExp achieves the highest guaranteed AUC (See Table \ref{tab:label_oodauc_dist}), showing the value of accounting for the noise distribution, instead of relying only on bounds on the input noise.

\section{Conclusion}
\label{sec:conclusion}
We have presented a general framework for verifying probabilistic specifications, and shown significant improvements upon existing methods, even for simple choices of the Lagrange multipliers where we can leverage efficient optimization algorithms.
We believe that our approach can be significantly extended by finding new choices of multipliers that capture interesting properties of the verification problem while leading to tractable optimization problems.
This could lead to discovery of new verification algorithms, and thus constitutes an exciting direction for future work.

\textbf{Limitations.}
We point out two limitations in the approach suggested by this work.
First, the guarantees provided by our approach heavily depend on the bounding method used to obtain the intermediate bounds -- this is consistent with prior work 
on verifying deterministic networks \citep{dathathri2020enabling, babnew}, where tighter bounds result in much stronger guarantees.
Second, as noted in Section \ref{sec:General_Functional}, 
our approach can only handle probability distributions that have bounded support, and alleviating this assumption would require further work.

\textbf{Broader Impact and Risks.}
Our work aims to improve the reliability of 
neural networks in settings where 
either the model or its inputs exhibit probabilistic behaviour. 
In this context, we anticipate this work to be largely beneficial and do not envision malicious usage.
However, the guarantees of our method crucially depend on having an accurate modeling of the uncertainty. 
Failing that can result in an overestimation of the reliability of the system, which can have catastrophic consequences in safety-critical scenarios.
In this regard, we advocate special care in the design of the specification when applying our approach to real-world use cases.
\newpage

\bibliographystyle{plainnat}
\bibliography{references}

\appendix
\onecolumn
\section*{Appendix}

\section{Probabilistic Specifications: Examples}
\label{supp:specform}
Below we provide further examples of specifications that can be captured by our framework.
\paragraph{Uncertainty calibration.}
Another desirable specification towards ensuring reliable uncertainty calibration for NNs is that the expected 
uncertainty in the predictions increases monotonically with an increase in the variance
of the input-noise distribution.
Formally, consider a set of zero-mean distributions $\mathcal{P}_{noise}$ with diagonal covariance matrices.
For any two such noise distributions $p_{\omega_1}, p_{\omega_2} \in \mathcal{P}_{noise}$ such that $\mathrm{Var}(p_{\omega_1}) \succcurlyeq \mathrm{Var}(p_{\omega_2})$ (where $\succcurlyeq$ is the element-wise inequality and  $\mathrm{Var}$ denotes the diagonal of the 
covariance matrix) and a given image $x$ from the training distribution, we wish to guarantee that 
the expected entropy of the resulting predictions corresponding to $p_{\omega_1}$ is greater than that of $p_{\omega_2}$, i.e., $\ExP{H(\texttt{softmax}(\phi(x_1)}{x_1 \sim x + p_{\omega_1}} \geq \ExP{H(\texttt{softmax}(\phi(x_2)}{x_2 \sim x + p_{\omega_2}}$, where $H$ is the entropy functional: $H(p) = 
- \overset{\lvert \mathcal{Y}  \rvert}{\underset{i=1}{\sum}} p_i \log p_i$.
Intuitively, this captures the desired behaviour that as the strength of the noise grows, 
the expected uncertainty in the network 
predictions increases.
We can capture this specification within the formulation described by equation \ref{eq:spec}, 
by letting:
\begin{enumerate}
    \item $\mathcal{P}_0 = \mathcal{P}_{noise} \times \mathcal{P}_{noise}$, 
    \item For a given NN $\phi$ and an input $x$, we define another network $\bar{\phi}: \mathcal{X} \times \mathcal{X} \rightarrow \mathcal{P}(\mathcal{Y}) \times \mathcal{P}(\mathcal{Y})$, where 
$\bar{\phi}$ is  such that:
$\bar{\phi}(a,b) = (\phi(x + a), \phi(x + b))$.
\end{enumerate}

We can then define the verification problem as certifying that the following holds :
\begin{align*}
\spec\br{\bar{\phi}{\br{\bar{p}_{0}, \bar{\bar{p}}_{0}}}}\coloneqq - \br{\ExP{H\br{\softmax{y_1}} - H\br{\softmax{y_2}}}{
\br{y_1, y_2} \sim \bar{\phi}\br{\bar{p}_{0}, \bar{\bar{p}}_{0}}
}} \leq 0,
\\
\forall \br{\bar{p}_{0}, \bar{\bar{p}}_{0}} \in \mathcal{P}_0 \textrm{ such that } \mathrm{Var}(\bar{p}_{0}) \succcurlyeq \mathrm{Var}(\bar{\bar{p}}_{0}). 
\end{align*}

\paragraph{Robust VAE}
In \cite{dathathri2020enabling}, the authors 
consider a specification that corresponds to certifying low reconstruction losses for a VAE 
decoder over a set of inputs in the 
neighborhood of the latent-variable mean predicted by the encoder for a given image. 
A natural generalization of this specification 
is one where low reconstruction error 
is guaranteed in expectation, since in practice 
the latent-representations that are fed into the decoder are drawn from a normal distribution whose mean and variance are predicted by the encoder.
A more general specification is one where we wish 
to verify that for a set of norm-bounded points around a given input, the expected reconstruction error from the VAE is small.
Formally, for a VAE $\phi$ (note that 
a VAE directly fits within our framework, where the distribution for the latent-variable 
can be obtained as the output of a stochastic 
layer), a given threshold $\tau \in \mathbb{R}_+$
and for a set of inputs $\mathcal{S}$,
we wish to certify that the following holds:
\begin{align}
\forall p_0 \in \mathcal{P}_0, \quad \spec(\phi(p_0)) \coloneqq \ExP{\lVert 
\ExP{s}{s \sim p_0}- {\mu}^{s'}_\phi \rVert_2^2 }{{\mu}^{s'}_\phi  \sim \phi(p_0)} - \tau \leq 0;
\label{eq:vae_spec}
\end{align}
where $\mathcal{P}_0 = \{\delta_s: s \in \mathcal{S}\}$.
\section{Proof of Functional Lagrangian Theorem} 
\label{app:funlag_thm}

\subsection*{Proof of Theorem \ref{thm:dual}}
\begin{proof}
The optimization problem \eqref{eq:specopt} is equivalent to the following optimization problem:
\begin{subequations}
\begin{align}
    \max_{p_0, p_1, \ldots, p_K} \quad  & \spec\br{p_K} \\
    \text{Subject to } & p_{k+1}\br{y} = \int \pi_k\br{y|x} p_k\br{x}dx \quad \forall y \in \X_{k+1}, \: \forall k \in \{0, \ldots, K-1\} \nonumber \\
    & p_0 \in \mathcal{P}_0
\end{align}\label{eq:orig_opt}
\end{subequations}
where $\spec$ is a functional of the output distribution $p_K$. We refer to the space of probability measures on $\X_k$ as $\mathcal{P}_k$ (not that for $k=0$, this may not be the whole space of probability measures but a constrained set of measures depending on the specification we would like to verify). The dual of this optimization problem can be written as follows:
\begin{align*}
    \max_{p_0 \in \mathcal{P}_0, p_1 \in \mathcal{P}_1, \ldots} & \spec\br{p_K}- \sum_{k=0}^{K-1} \int \lambda_{k+1}\br{z}p_{k+1}\br{z}dz \\
    & + \sum_{k=0}^{K-1} \displaystyle\int_{\X_{k+1}, \X_k} \lambda_{k+1}\br{z}\pi_k\br{z|x}p_k\br{x}dx dz, 
\end{align*}
where we assigned Lagrange multipliers $\lambda_{k+1}\br{y}$ for every $y \in \X_{k+1}$. The above optimization problem can be decomposed as follows:
\begin{align*}
& \spec\br{p_K}-\int_{\X_K} \lambda_K\br{x}p_K\br{x}dx  \\ & + \sum_{k=0}^{K-1} \int_{\X_{k}} \br{\int_{\X_{k+1}} \lambda_{k+1}\br{y}\pi_k\br{y|x}dy - \lambda_{k}\br{x}} p_k\br{x}dx.
\end{align*}
This is a separable problem in each $p_k$ and since $p_k$ is constrained to be a probability measure, the optimal choice of $p_k$ (for $k=1, \ldots, K-1$) is a $\delta$ measure with probability $1$ assigned to the the $x \in \X_k$ that maximizes:
\[\int_{\X_{k+1}} \lambda_{k+1}\br{y}\pi_k\br{y|x}dy - \lambda_{k}\br{x}.\]

The optimization over $p_0$ can be rewritten as follows:
\[\max_{p_0 \in \mathcal{P}_0} \int_{\X_0}\br{\int_{\X_1} \lambda_1\br{y}\pi_0\br{y|x}dy}p_0\br{x}dx.\]

The optimization over $p_K$ can be rewritten as follows:
\[\spec^\star\br{\lambda_K}=\max_{p_K \in \mathcal{P}_K} \spec\br{p_K}-\int \lambda_K\br{x}p_K\br{x}dx.\]

The overall dual problem can be rewritten as:
\begin{align*}
&     \spec^\star\br{\lambda_K} + \sum_{k=1}^{K-1} \max_{x \in \X_k}\br{\int_{\X_{k+1}} \lambda_{k+1}\br{y}\pi_k\br{y|x}dy - \lambda_{k}\br{x}} \\
& \quad + \max_{p_0 \in \mathcal{P}_0} \int_{\X_0}\br{\int_{\X_1} \lambda_1\br{y}\pi_0\br{y|x}dy}p_0\br{x}dx.
\end{align*}

Writing this in terms of expected values, we obtain $g\br{\lambda}$. 
Plugging in $\lambda^\star$ into $g\br{\lambda}$, all the terms cancel except the first term which evaluates to:
\begin{equation*}
\max_{p_0 \in \Pcal_0} \ExP{\ExP{\lambda_1^\star\br{x}}{y \in \pi_0\br{x}}}{x \sim p_0}
= \max_{p_0 \in \Pcal_0} \ExP{\ExP{\ExP{\lambda_2^\star\br{z}}{z \sim \pi_1\br{x}}}{y \in \pi_0\br{x}}}{x \sim p_0}
\ldots = \max_{p_0 \in \Pcal_0} \spec\br{p_0}
\end{equation*}

\end{proof}

\section{Additional Theoretical Results}
\label{app:additional_theory}

\subsection{Computation of Expected Values} 
\label{app:appendix_expectation}
Since $\pi_k\br{x}$ is typically a distribution that one can sample from easily (as it is required to perform forward inference through the neural network), estimating this expectation via sampling is a viable option. However, in order to turn this into verified bounds on the specification, one needs to appeal to concentration inequalities and the final guarantees would only be probabilistically valid. We leave this direction for future work.

Instead, we focus on situations where the expectations can be computed in closed form. 
In particular, we consider layers of the form
$\pi_k\br{x} = w s\br{x} + b, (w, b) \sim \pw_k$, where $s$ is an element-wise function like \text{ReLU}, sigmoid or tanh and $(w, b)$ represents a fully connected or convolutional layer. 
We consider a general form of Lagrange multipliers as a sum of quadratic and exponential terms as follows:
\[\lambda\br{x} = q^T x + \frac{1}{2} x^T Q x + \kappa \exp\br{\gamma^T x}.\]
Let:
\[\tilde{s}\br{x} = \begin{pmatrix} 1 \\ s\br{x} \end{pmatrix}, \tilde{Q} = \begin{pmatrix} 0 & q^T \\ q & Q\end{pmatrix}, \tilde{w} = \begin{pmatrix} b & w \end{pmatrix}.\]
Then: 
\[\lambda\br{Ws\br{x}+b}=\frac{1}{2} \br{\tilde{s}\br{x}}^T \tilde{w}^T\tilde{Q}\tilde{w} \tilde{s}\br{x} + \kappa \exp\br{\gamma^T\tilde{w}\tilde{s}\br{x}}.\]
Taking expected values with respect to $\tilde{W} \sim \pw_k$, we obtain:
\begin{align}
&\ExP{\lambda\br{ws\br{x}+b}}{} =  \frac{1}{2}\br{\tilde{s}\br{x}}^T\ExP{\tilde{w}^T\tilde{Q}\tilde{w}}{\tilde{s}\br{x}} + \kappa \prod_{i, j} \ExP{\exp\br{\gamma_i \tilde{w}_{ij}\tilde{s}\br{x_j}}}{},    \label{eq:lamquad_exp}
\end{align}
where we have assumed that each element of $\tilde{W}_{ij}$ is independently distributed.
The first term in \eqref{eq:lamquad_exp} evaluates to: 
\[\frac{1}{2}\text{Trace}\br{\text{Cov}\br{\tilde{w}\tilde{s}\br{x}}\tilde{Q}} + \frac{1}{2} \br{\ExP{\tilde{w}\tilde{s}\br{x}}{}}^\top\tilde{Q} \br{\ExP{\tilde{w}\tilde{s}\br{x}}{}},\]
and the second one to:
\[\kappa \prod_{i, j} \text{mgf}_{ij}\br{\gamma_i \tilde{s}_j\br{x}},\]
where $\text{Cov}\br{X}$ refers to the covariance matrix of the random vector, and $\text{mgf}_{ij}$ refers to the moment generating function of the random variable $\tilde{w}_{ij}$:
\[\text{mgf}_{ij}\br{\theta}=\ExP{\exp\br{\tilde{w}_{ij}\theta}}{}.\]

 The details of this computation for various distributions on $\tilde{w}$ (Gaussian posterior, MC-dropout) are worked out below.
 
 \paragraph{Diagonal Gaussian posterior.} 
 Consider a BNN with a Gaussian posterior, $\tilde{w} \sim \N\br{\mu, \diagmat{\sigma^2}}$, where $\mu, \sigma \in \R^{mn}$, let $\text{Mat}\br{\mu} \in \R^{m \times n}$ denote a reshaped version of $\mu$. Then, we have:
 \begin{align*}
\ExP{\lambda\br{ws\br{x}+b}}{} 
    &= \frac{1}{2}\text{Trace}\br{\diagmat{\text{Mat}\br{\sigma^2}\tilde{s}\br{x}}\tilde{Q}} + \frac{1}{2} {\br{\text{Mat}\br{\mu}\tilde{s}\br{x}}}^T \tilde{Q}\br{\text{Mat}\br{\mu}\tilde{s}\br{x}} \\
    &+ \kappa \prod_{i, j} \exp\br{\text{Mat}\br{\mu}_{ij}\tilde{s}\br{x_j}\gamma_i+\frac{1}{2}\text{Mat}\br{\sigma^2}_{ij}\br{\tilde{s}\br{x_j}\gamma_i}^2}.     
 \end{align*}
 
\paragraph{MC dropout.} 
Now assume a neural network with dropout: $\tilde{w} = \mu \odot \text{Bernoulli}\br{\pdrop}$, where $\mu \in \R^{mn}$ denotes the weight in the absence of dropout and $\pdrop\in\R^{mn}$ denotes the probability of dropout. 
Then, we have:
 \begin{align*}
 \ExP{\lambda\br{ws\br{x}+b}}{}  
    &=\frac{1}{2}\text{Trace}\br{\diagmat{\text{Mat}\br{\mu \odot \pdrop \odot \br{1-\pdrop}}\tilde{s}\br{x}}\tilde{Q}} \\
    &+ \frac{1}{2} {\br{\text{Mat}\br{\mu \odot \pdrop}\tilde{s}\br{x}}}^T \tilde{Q}\br{\text{Mat}\br{\mu \odot \pdrop}\tilde{s}\br{x}} \\
    &+ \kappa \prod_{i, j} \br{\text{Mat}\br{\pdrop}_{ij} \exp\br{\text{Mat}\br{\mu}_{ij}\tilde{s}\br{x_j}\gamma_i}+   1-\text{Mat}\br{\pdrop}_{ij}}.   
 \end{align*}

\subsection{Expected-Softmax Optimization}
\label{app:softmax_opt}
We describe an algorithm to solve optimization problems of the form
\[\max_{\ell \leq x \leq u} \mu^T \softmax{x} - \lambda ^ T x\]
Our results will rely on the following lemma:
\begin{proposition}\label{thm:softmax_exact_prop}
Consider the function
\[f\br{x}= \frac{\sum_i \mu_i\exp\br{x_i}+D}{\sum_j \exp\br{x_j} + B} - \lambda ^ T x\]
where $B \geq 0$ and $D=0$ if $B=0$. Let $r=\frac{D}{B}$ if $B>0$ and $0$ otherwise. Define the set
\[\Delta = \left\{\kappa \in \R: \br{\kappa-r} \br{\prod_{i=1}^n \br{\mu_i-\kappa}} -\sum_{i=1}^n \mu_i \br{1-r}\lambda_i \br{\prod_{j \neq i}  \br{\mu_j-\kappa}}=0\right\}\]
which is a set with at most $n+1$ elements. Define further 
\[\Delta_f = \begin{cases}\left\{\kappa \in \Delta: 0 < \frac{\lambda}{\mu-\kappa} < 1, \sum_{i=1}^n \frac{\lambda_i}{\mu_i-\kappa} \leq 1\right\} & \text{ if } B > 0 \\
\left\{\kappa \in \Delta: 0 < \frac{\lambda}{\mu-\kappa} < 1, \sum_{i=1}^n \frac{\lambda_i}{\mu_i-\kappa} = 1\right\} & \text{ if } B = 0
\end{cases}
\]
Then, the set of stationary points of $f$ are given by
\[\left\{\log\br{h\br{\frac{\lambda}{\mu-\kappa}}}: \kappa \in \Delta_f\right\}\]
where 
\[h\br{v} = \begin{cases}
\frac{B v}{1-\One^T v} \text{ if } B > 0 \\
\{\theta v: \theta > 0\} \text{ if } B=0
\end{cases}
\]
\end{proposition}
\begin{proof}
Differentiating with respect to $x_i$, we obtain
\[ \frac{\exp\br{x_i}}{\sum_j \exp\br{x_j}+B}\br{\mu_i - \br{\frac{\sum_j \mu_j \exp\br{x_j}+D}{\sum_j \exp\br{x_j} + B}}} - \lambda_i = p_i\br{\mu_i -\mu^T p-q} - \lambda_i\]
where
\[p_i = \frac{\exp\br{x_i}}{\sum_j \exp\br{x_j} + B}, q=\frac{D}{\sum_j \exp\br{x_j} + B}.\]
If we set the derivative to $0$ (to obtain a stationary point)
we obtain the following coupled set of equations in $p, q, \kappa$:
\begin{align*}
    p_i &= \frac{\lambda_i}{\mu_i - \kappa} \quad i=1, \ldots, n \\
    q &= r\br{1-\sum_i p_i}, \\
    \kappa &= \sum_i \mu_i p_i + q,
\end{align*}

where $r=\frac{D}{B}$. 
We can solve this by first solving the scalar equation
\[\kappa - r= \sum_i \frac{\mu_i\br{1-r} \lambda_i}{\mu_i - \kappa}\]
for $\kappa$ (this is derived by adding up the first $n$ equations above weighted by $\mu_i$ and plugging in the value of $q$). This equation can be converted into a polynomial equation in $\kappa$
\[\br{\kappa-r} \br{\prod_i \br{\mu_i-\kappa}} -\sum_i \mu_i\br{1-r} \lambda_i \br{\prod_{j \neq i}  \br{\mu_j-\kappa}}=0\]
which we can solve for all possible real solutions, denote this set $\Delta$. Note that this set has at most $n+1$ elements since it is the set of real solutions to a degree $n+1$ polynomial.

In order to recover $x$ from this, we first recall:
\[p_i = \frac{\lambda_i}{
\mu_i-
\kappa}\]
Since $p_i=\frac{\exp{x_i}}{\sum_j \exp\br{x_j} + B}$, we require that $p_i \in (0, 1)$ and $\sum_i p_i \leq 1$ (with equality when $B=0$ and strict inequality when $B=1$. We thus filter $\Delta$ to the set of $\kappa$ that lead to $p$ satisfying these properties to obtain $\Delta_f$. 

Once we have these, we are guaranteed that for each $\kappa \in \Delta_f$, we can define $p_i$ as above and solve for $x_i$ by solving the linear system of equations
\[u_i=p_i\br{\sum_j u_j+B} \quad i=1,2,\ldots, n\]
which can be solved as:
\[u = B\br{I-p\One^T}^{-1}p=\frac{B p}{1-\One^T p}, x=\log\br{u}\] 
if $B > 0$ since the matrix $I-p\One^T$ is strictly diagonally dominant and hence invertible, and we applied the Woodbury identity to compute the explicit inverse.  

If $B=0$, we have $p=\softmax{x}$ and can recover $x$ as
\[x=\log\br{p\theta}\] for any $\theta > 0$.
\end{proof}

The above lemma allows us to characterize all stationary points of the function
\[\mu^T \softmax{x} - \lambda^T x\]
when a subset of entries of $x$ are fixed to their upper or lower bounds, and we search for stationary points wrt the remaining free variables. Given this ability, we can develop an algorithm to globally optimize $\mu^T \softmax{x} - \lambda^T x$ subject to bound constraints by iterating over all possible $3^n$ configurations of binding constraints (each variable could be at its lower bound, upper bound or strictly between them). In this way, we are guaranteed to loop over all local optima, and by picking the one achieving the best objective value, we can guarantee that we have obtained the global optimum. The overall algorithm is presented in Algorithm \ref{alg:exhaustive_softmax}.

\begin{proposition}
Algorithm \ref{alg:exhaustive_softmax} finds the global optimum of the optimization problem
\[\min_{x: \ell \leq x \leq u} \mu^T \softmax{x}-\lambda^T x\]
and runs in time $O(n 3^n)$ where $n$ is the dimension of $x$.
\end{proposition}

\begin{algorithm}[htb]
   \caption{Solving softmax layer problem via exhaustive enumeration}
   \label{alg:exhaustive_softmax}
\begin{algorithmic}
\STATE Inputs: $\lambda, \mu, \ell, u \in \R^n$
\STATE $x^\star \gets \ell$
\STATE $f_{opt}\br{x}\gets \mu^T \softmax{x}-\lambda^T x$, $f^\star \gets f_{opt}\br{x^\star}$
   \FOR{$v \in {[\text{Lower}, \text{Upper}, \text{Interior}]}^n$}
   \STATE $\text{nonbinding}[i]\gets \br{v[i]==\text{Interior}}$, $x_i\gets\begin{cases} l[i] \text{ if } v[i]=\text{Lower} \\ u[i] \text{ if } v[i]=\text{Upper} \end{cases}$ for $i=1, \ldots, n$
       \STATE $B \gets \displaystyle\sum_{i \text{ such that } \text{nonbinding}[i]==\text{False}} \exp\br{x[i]}$
       \STATE $D \gets \displaystyle\sum_{i \text{ such that } \text{nonbinding}[i]==\text{False}} \mu[i]\exp\br{x[i]}$

    \STATE Use proposition \ref{thm:softmax_exact_prop} to find the set of stationary points $\mathcal{S}_f$ of the function
    \[f\br{x}=\frac{\displaystyle\sum_{i \text{ such that } \text{nonbinding}[i]} \mu_i\exp\br{x_i}+D}{\displaystyle\sum_{j \text{ such that } \text{nonbinding}[j]} \exp\br{x_j} + B}-\sum_{j \text{ such that } \text{nonbinding}[j]} \lambda[j]x_j\]
    \FOR{$x^s \in \mathcal{S}_f$}
    \IF{$x^s_i \in [\ell[i], u[i]] \quad \forall i\text{ s.t } \text{nonbinding}[i]$}
      \STATE $x_i\gets x^s_i \quad \forall i \text{ s.t } \text{nonbinding}[i]$.
    \IF{$f_{opt}\br{x} > f^\star$}
    \STATE $x^\star \gets x$
    \STATE $f^\star \gets f_{opt}\br{x^\star}$
    \ENDIF
    \ENDIF
    \ENDFOR
    \ENDFOR
\STATE Return $x^\star, f^\star$
\end{algorithmic}
\end{algorithm}

\newpage

\subsection{Input Layer with Linear-Exponential Multipliers} \label{app:LinExP}
We recall the setting from Proposition \ref{prop:linexp}. 
Let $\lambda_1\br{x} = \alpha^T x + \exp\br{\gamma^T x+\kappa}$, $\lambda_2\br{x}=\beta^T x$, $g_0^\star = \max_{p_0 \in \Pcal_0} g_0\br{p_0, \lambda_1}$, and $g_1^\star=\max_{x \in \X_1} g_0\br{x, \lambda_1, \lambda_2}$.
\begin{proposition}
In the setting described above, and with $s$ as the element-wise activation function:
\begin{align*}
& g_0^\star \leq  \alpha^T\br{ w \mu + b} +  \exp\br{\frac{\norm{w^T\gamma}^2\sigma^2}{2} + \gamma^T b + \kappa}, \\
& g_1^\star\leq \max_{\substack{x \in \X_2 \\ z = s\br{x}}} \beta^T \br{w_2z+b_2} -\alpha^Tx -\exp\br{\gamma^T x + \kappa}.
\end{align*}
The maximization in the second equation can be bounded by solving the following convex optimization problem:
\begin{align*}
 \min_{\eta \in \R^n, \zeta \in \R_+} &\zeta \br{\log\br{\zeta}-1-\kappa} + \One^T\max \br{\br{\eta+w_2^T\beta} \odot s\br{l_2}, \br{\eta+w_2^T\beta} \odot s\br{u_2}} \\
&+ \sum_i s^\star\br{\alpha_i+\zeta\gamma_i, \eta_i, l_{2i}, u_{2i}},
\end{align*}
where $s^\star\br{a, b, c, d} = \max_{z \in [c, d]} -az-bs\br{z}$.
\end{proposition}
\begin{proof}
\begin{align*}
\max_{\substack{x \in \X_2 \\ z = s\br{x}}} &\beta^T \br{w_2z+b_2} -\alpha^Tx -\exp\br{\alpha^T x + \kappa}, \\
    &\leq \min_{\eta} \max_{x \in \X_2, z \in s\br{\X_2}} \eta^T\br{z-s\br{x}}+\beta^T \br{w_2z+b_2} -\alpha^Tx -\exp\br{\gamma^T x + \kappa}, \\
    &\leq \min_{\eta, \zeta} \max_{x \in \X_2, t} \One^T\max \br{\br{\eta+w_2^T\beta} \odot s\br{l_2}, \br{\eta+w_2^T\beta} \odot s\br{u_2}}+\beta^T b_2 -\alpha^Tx \\
    &\quad - \eta^T s\br{x} -\exp\br{t} + \zeta\br{t-\gamma^T x - \kappa}, \\
    &\leq \min_{\eta, \zeta} \zeta \br{\log\br{\zeta}-1-\kappa} + \One^T\max \br{\br{\eta+w_2^T\beta} \odot s\br{l_2}, \br{\eta+w_2^T\beta} \odot s\br{u_2}} \\
    &\quad + \max_{l_2 \leq x \leq u_2} -\br{\alpha + \zeta \gamma}^T x - \eta^T s\br{x}, \\
    &\leq \min_{\eta, \zeta} \zeta \br{\log\br{\zeta}-1-\kappa} + \One^T\max \br{\br{\eta+w_2^T\beta} \odot s\br{l_2}, \br{\eta+w_2^T\beta} \odot s\br{u_2}} \\
    &\quad + \sum_i \max_{l_{2i} \leq x_i \leq u_{2i}} -\br{\alpha_i + \zeta \gamma_i} x_i - \eta_i s\br{x_i}. 
\end{align*}
\end{proof}

\subsection{Inner Problem with Linear Multipliers}
\label{app:linearmults}

In its general form, the objective function of the inner maximization problem can be expressed as:
\begin{equation}
g_k(x_k,\lambda_k, \lambda_{k+1}) = \ExP{\lambda_{k+1}\br{y}}{y \sim \pi_k\br{x_k}}-\lambda_{k}\br{x_k}.    
\end{equation}
We assume that the layer is in the form $y = W s\br{x} + b$, where $W$ and $b$ are random variables and $s$ is an element-wise activation function.
Then we can rewrite $g_k(x_k,\lambda_k, \lambda_{k+1})$ as:
\begin{equation}
g_k(x_k,\lambda_k, \lambda_{k+1}) = \ExP{\lambda_{k+1}\br{W \max\{x_k, 0\} + b}}{W, b}-\lambda_{k}\br{x_k}.
\end{equation}
We now use the assumption that $\lambda_{k+1}$ is linear: $\lambda_{k+1}: y \mapsto \theta_{k+1}^\top y$. Then the problem can equivalently be written as:
\begin{equation}
\begin{split}
g_k(x_k,\lambda_k, \lambda_{k+1}) 
    &= \ExP{\theta_{k+1}^\top\br{W s\br{x_k} + b}}{W, b}-\theta_k^T x_k, \\ 
    &= \br{\ExP{W}{}^\top\theta_{k+1}}^\top s\br{x_k}+ \theta_{k+1}^\top \ExP{b}{} -\theta_{k}^Tx_k, \\ 
    &= \theta_{k+1}^T\ExP{b}{}+\sum_i \br{\ExP{W}{}^\top\theta_{k+1}}_i s\br{x_i}-\br{\theta_k}_i x_i.
\end{split}
\end{equation}
Maximizing the RHS subject to $l \leq x \leq u$, we obtain:
\[\theta_{k+1}^T\ExP{b}{}+\sum_i \max_{z \in [l_i, u_i]}\br{\ExP{W}{}^\top\theta_{k+1}}_i s\br{z}-\br{\theta_k}_i z.\]
where the maximization over $z$ can be solved in closed form for most common activation functions $s$ as shown in \citet{dvijotham2018dual}.

So we can simply apply the deterministic algorithm to compute the closed-form solution of this problem.

\section{Relationship to Prior work} \label{app:relationship}
We establish connections between the functional Lagrangian framework and prior work on deterministic verification techniques based on Lagrangian relaxations and SDP relaxations.

\subsection{Lagrangian Dual Approach}
\label{app:dvijeq}
We assume that the network layers are deterministic layers of the form:
\begin{equation}
\begin{split} \label{eq:det_network_layers}
  \forall \: k, k \text{ is odd } \: \pi_k\br{x} &= w_kx+b_k, \\
  \forall \: k, k \text{ is even } \: \pi_k\br{x} &= s\br{x},
\end{split}
\end{equation}
where $s$ is an element-wise activation function 
and that the specification can be written as:
\begin{equation}  \label{eq:det_network_obj}
\spec\br{x_K}=c^T x_K.    
\end{equation}

\begin{proposition}[Linear Multipliers]
For a verification problem described by equations (\ref{eq:det_network_layers}, \ref{eq:det_network_obj}), the functional Lagrangian framework with linear functional multipliers $\lambda_k\br{x}=\theta_k^T x$ is equivalent to the Lagrangian dual approach from \cite{dvijotham2018dual}.
\label{thm:linmult}
\end{proposition}
\begin{proof}
The final layer problem is
\[\max_{x_K} c^Tx_K-\theta_K^T x_K = \One^T\max\br{\br{c-\theta_K} \odot l_K, \br{c-\theta_K} \odot u_K}\]
For even layers with $k < K$, the optimization problem is
\begin{align*}
& \max_{x \in [l_k, u_k]} \theta_{k+1}^T\br{w_kx+b_k}-\theta_k^T x = \theta_{k+1}^Tb_k + \br{w_k^T\theta_{k+1}-\theta_k}^T x \\
& = \One^T \max\br{\br{w_k^T\theta_{k+1}-\theta_k} \odot l_k, \br{w_k^T\theta_{k+1}-\theta_k} \odot u_k} + \theta_{k+1}^Tb_k
\end{align*}
For odd layers with $k < K$, the optimization problem is
\begin{align*}
\max_{x \in [l_k, u_k]} \theta_{k+1}^Ts\br{x}-\theta_k^T x
= \sum_i \max_{z \in [l_{ki}, u_{ki}]} \br{\theta_{k+1}}_i s\br{z}-\br{\theta_k}_i z
\end{align*}
All these computations precisely match those from \citet{dvijotham2018dual}, demonstrating the equivalence.
\end{proof}

\subsection{SDP-cert}
\label{app:sdpcert}
We assume that the network layers are deterministic layers of the form:
\begin{equation}
\begin{split} \label{eq:det_network_layers_sdp}
  \forall \: k \: \pi_k\br{x} &= \relu\br{w_kx+b_k}
\end{split}
\end{equation}
where $s$ is an element-wise activation function 
and that the specification can be written as:
\begin{equation}  \label{eq:det_network_obj_sdp}
\spec\br{x_K}=c^T x_K.    
\end{equation}

\begin{proposition}[Quadratic Multipliers]
For a verification problem described by equations (\ref{eq:det_network_layers_sdp}, \ref{eq:det_network_obj_sdp}), the optimal value of the Functional Lagrangian Dual with
\begin{align*}
& \lambda_k\br{x}=q_k^T x + \frac{1}{2} x^T Q_k x \quad k=1, \ldots, K-1\\
& \lambda_K\br{x}=q_K^T x
\end{align*}
and when an SDP relaxation is used to upper bound the inner maximization problems over $g_k$, is equal to the dual of the SDP relaxation from \cite{raghunathan2018semidefinite}.
\end{proposition}
\begin{proof}
\newcommand{\qp}{\tilde{q}}
\newcommand{\lp}{\tilde{l}}
\newcommand{\up}{\tilde{u}}
\newcommand{\Qp}{\tilde{Q}}
With quadratic multipliers of the form $\lambda_k\br{x} = q_k^T x + \frac{1}{2}x^T Q_k x \quad k=1, \ldots, K-2$ and  $\lambda_K\br{x} = q_K^T x$, the inner maximization problems for the intermediate layers are of the form:
\[\max_{\substack{x \in [l, u] \\ y =\relu\br{wx+b}}} \qp^T y + \frac{1}{2}y^T \Qp y-q^T x - \frac{1}{2} x^T Q x,\]
where $l=l_k, u=u_k, \qp=q_{k+1}, \Qp=Q_{k+1}, q=q_k, Q=Q_k, w=w_k, b=b_k$. Let $x \in \R^n, y \in \R^m$ ($n$ dimensional input, $m$ dimensional output of the layer). Further, let $\tilde{l}=l_{k+1}, \tilde{u}=u_{k+1}$.

We can relax the above optimization problem to the following Semidefinite Program (SDP) (following \citet{raghunathan2018semidefinite}):

\begin{subequations}\label{eq:Popt_SDP}
\begin{align}
\max_{P} &\qp^T P[y] + \frac{1}{2}\text{Trace}\br{\Qp P[yy^T]} -q^T P[x] - \frac{1}{2} \text{Trace}\br{Q P[xx^T]} \\
\text{Subject to } & P = \begin{pmatrix} 1 & \br{P[y]}^T & \br{P[x]}^T \\ P[y] & P[yy^T] & P[xy^T] \\
P[x] & \br{P[xy^T]}^T & P[xx^T]\end{pmatrix}    \in \Sym^{n+m+1}, \\
& P \succeq 0, \\
& \diagmat{P[xx^T]-l\br{P[x]}^T-P[x]u^T+lu^T} \leq 0, \\
& \diagmat{P[yy^T]-\lp \br{P[y]}^T-P[y]\up^T+\lp \up^T} \leq 0, \\
& P[y] \geq 0, P[w] \geq wP[x], \\
& \diagmat{wP[xy^T]}+P[y]\odot b=\diagmat{P[yy^T]}.
\end{align}
\end{subequations}
where the final constraint follows from the observation that $y \odot \br{y-wx-b} = 0$.

The above optimization problem resembles the formulation of \citet{raghunathan2018semidefinite} except that it only involves two adjacent layers rather than all the layers at once. Let $\Delta_k$ denote the feasible set given the constraints in the above optimization problem. Then, the formulation of \citet{raghunathan2018semidefinite} can be written as:
\begin{subequations}\label{eq:SDP}
\begin{align}
\max & \,\, c^T y_K    \\
\text{subject to } & P_{k} \in \Delta_k \quad k=0, \ldots, K-1, \,\, l_K \leq y_K \leq u_K, \\
& P_{k+1}[xx^T]=P_k[yy^T] \quad k=0, \ldots, K-2, \\
& P_{k+1}[x]=P_k[y]\quad k=0, \ldots, K-2, \\
& y_K = P_{K-1}[y].
\end{align}
\end{subequations}
Note that in \citet{raghunathan2018semidefinite}, a single large $P$ matrix is used whose block-diagonal sub-blocks are $P_k$ and the constraint $P \succeq 0$ is enforced. Due to the matrix completion theorem for SDPs \citep{grone1984positive, vandenberghe2015chordal}, it suffices to ensure postitive semidefiniteness of the sub-blocks rather than the full $P$ matrix.

Dualizing the last three sets of constraints above with Lagrangian multipliers $\Theta_k \in \Sym^{n_k+n_{k+1}+1}, \theta_k \in \R^{n_k}$ and $\theta_K \in \R^{n_K}$, we obtain the following optimization problem:
\begin{align*}
\max & \,\, c^T y_K +\theta_K^T\br{-P_{K-1}[y]+y_K} + \sum_{k=0}^{K-2}\text{Trace}\br{\Theta_k\br{P_{k+1}[xx^T]-P_{k}[yy^T]}} \\
&+ \sum_{k=0}^{K-2}\theta_k^T\br{P_{k+1}[x]-P_{k}[y]}    \\
\text{subject to } & P_{k} \in \Delta_k \quad k=0, \ldots, K-1, \\
& l_K \leq y_K \leq u_K.
\end{align*}
The objective decomposes over $P_k$ and can be rewritten as:
\begin{equation}\label{eq:dual_fin}
\begin{split}
\max_{l_K \leq y_K \leq u_K} &\br{c+\theta_K}^T y_K + \sum_{k=0}^{K-1}\max_{P_k \in \Delta_k} \text{Trace}\br{\Theta_{k-1}P_{k}[xx^T]} + \theta_{k-1}^TP_k[x]-\theta_k^TP_k[y]\\
&-\text{Trace}\br{\Theta_{k}P_{k}[yy^T]}, 
\end{split}
\end{equation}
with the convention that $\Theta_{K-1}=0, \Theta_{-1}=0, \theta_{-1}=0$. If we set $Q_k=-\Theta_{k-1}, q_k=-\theta_{k-1}$ for $k=1, \ldots, K$, then the optimization over $P_k$ precisely matches the optimization in \eqref{eq:Popt_SDP}. Further, since $\lambda_K$ is linear, the final layer optimization simply reduces to:
\[\max_{l_K \leq x_K \leq u_K} c^Tx_K-q_K^Tx_K,\]
which matches the first term in \eqref{eq:dual_fin}.

Thus, the functional Lagrangian framework with quadratic multipliers $\lambda_k$ for $k=1, \ldots, K-2$ and a linear multiplier for $\lambda_K$ precisely matches the Lagrangian dual of \eqref{eq:SDP} and since \eqref{eq:SDP} is a convex optimization problem, strong duality guarantees that the optimal values must coincide. 

\end{proof}

\section{Additional Experimental Details}
\label{app:experimental_details}

\subsection{Robust OOD Detection on Stochastic Neural Networks}
\label{app:experimental_details_ood}

\paragraph{Inner Optimization.} 
All inner problems have a closed-form as shown in section \ref{app:linearmults}, except for the last one, which is handled as follows.

The last inner problem can be formulated as:
\begin{equation}
    \max_{x_K \in \X_K} \mu^\top \texttt{softmax}(x_K) + \nu^\top x_K,
    \label{eq:mlps}
\end{equation}
where $\mu$ is a one-hot encoded vector and $\nu$ is a real-valued vector.

\begin{itemize}
\item Projected Gradient Ascent (Training):
\begin{itemize}
    \item Hyper-parameters: we use the Adam optimizer \cite{kingma2014adam}, with a learning-rate of 1.0 and a maximum of 1000 iterations.
    \item Stopping criterion: when all coordinates have either zero gradient, or are at a boundary with the gradient pointing outwards of the feasible set.
    \item In order to help the gradient method find the global maximum, we use a heuristic for initialization, which consists of using the following two starting points for the maximization (and then to take the best of the corresponding two solutions found):
    \begin{enumerate}
        \item Ignore affine part ($\nu=0$), which gives a solution in closed form: set $x_K$ at its upper bound at the coordinate where $\mu$ is 1, and at its lower bound elsewhere.
        \item Ignore softmax part ($\mu=0$), which also gives a solution in closed form: set $x_K$ at its upper bound at the coordinates where $\nu \geq 0$, and at its lower bound elsewhere.
    \end{enumerate}
\end{itemize}

\item Evaluation: we use Algorithm \ref{alg:exhaustive_softmax} at evaluation time, which solves the maximization exactly.
\end{itemize}

\paragraph{Outer Optimization.} We use the Adam optimizer, with a learning-rate that is initialized at 0.001 and divided by 10 every 250 steps. We run the optimization for a total of 1000 steps.

\paragraph{Gaussian-MLP.} We use the ReLU MLP from \citep{wicker2020probabilistic} that consists of 2 hidden layers of 128 units each. The models are available at \url{https://github.com/matthewwicker/ProbabilisticSafetyforBNNs}.

\paragraph{LeNet.} We use the LeNet5 architecture with dropout applied to the last fully connected layer with a probability of 0.5. To make the bound-propagation simpler, we do not use max-pooling layers and instead increase the stride of convolutions.

\paragraph{VGG-X.} For VGG-X (where X $\in \{2, 4, 8, 16, 32, 64\}$), the architecture can be described as:
\begin{itemize}
    \item Conv 3x3, X filters, stride 1
    \item ReLU
    \item Conv 3x3, X filters, stride 2
    \item ReLU
    \item Conv 3x3, 2X filters, stride 2
    \item ReLU
    \item Conv 3x3, 2X filters, stride 2
    \item ReLU
    \item Flatten
    \item Linear with 128 output neurons
    \item Dropout with rate 0.2
    \item Linear with 10 output neurons
\end{itemize}

\paragraph{Hardware}
The verification of each sample is run on a CPU with 1-2 cores (and on each instance, BP and FL are timed on the same exact hardware configuration).
\subsection{Adversarial Robustness for Stochastic Neural Networks}
\label{app:experimental_details_adv}
\paragraph{Inner Optimization.}
We use a similar approach as in Appendix \ref{app:experimental_details_ood}.
For the final inner problem (corresponding to the objective which is a linear function of the softmax and the layer inputs), we run projected gradient ascent during the 
optimization phase and then use Algorithm \ref{alg:exhaustive_softmax} to solve the maximization exactly.
For projected gradient ascent, because of the 
non-convexity of the problem, 
we use the following heuristics to try and find the global maximum:
\begin{itemize}
    \item Black-box attack (1st phase): we use the \texttt{Square} adversarial attack \cite{ACFH2020square}, with 600 iterations, 300 random restarts and learning-rate of 0.1.
    \item Fine-tuning (2nd phase): We then choose the best attack from the restarts, and employ projected gradient ascent, with a learning-rate of 0.1 and 100 iterations to fine-tune further.
\end{itemize}

\paragraph{Model Parameters.} We use the 1 and 2 layer ReLU MLPs from \citep{wicker2020probabilistic}. The models are available at \url{https://github.com/matthewwicker/ProbabilisticSafetyforBNNs}.

\paragraph{Outer Optimization.} 
We use the Adam optimizer, with a learning-rate that is initialized at 0.001 and divided by 10 every 1000 steps. We run the optimization for a total of 3000 steps.
\paragraph{Hardware}
All experiments were run on a \textsc{P100 GPU}.
\subsection{Distributionally Robust OOD Detection}
\label{app:experimental_details_dist_ood}

\paragraph{Model.}
We train networks on MNIST
using the code from \url{https://gitlab.com/Bitterwolf/GOOD} with the CEDA 
method, and with the default hyperparameters.
We train a CNN with \texttt{ReLU} activations the following layers:
\begin{itemize}
    \item Conv 4x4, 16 filters, stride 2, padding 2 on both sides
    \item ReLU
    \item Conv 4x4, 32 filters, stride 1, padding 1 on both sides
    \item Relu
    \item Flatten
    \item Linear with 100 output neurons
    \item Relu
    \item Linear with 10 output neurons
\end{itemize}

\paragraph{Outer Optimization.}
For the outer loop of the verification procedure, we use Adam for 100k steps.
The learning-rate is initially set to 0.0001 and then divided by 10 after 60k and 80k steps.

\paragraph{Hardware}
We run the experiments for this section on a CPU with 2-4 cores.
\section{Additional Results with Interval Bound Propagation for Bilinear Operations}

\subsection{Robust OOD Detection for Stochastic Neural Networks}
We repeat the experiments in Section \ref{sec:robustood} where we use IBP to handle bound-propagation through the layers where bilinear propagation is required (because of bounds coming from both the layer inputs and the layer parameters due to the stochasticity of the model) instead of \citet{bunel2020efficient}.
\citet{bunel2020efficient} usually results in significantly tighter bounds compared to IBP but we note that for MNIST-CNN and CIFAR-CNN, we expect IBP to perform competitively as the bilinear bound propagation is only applied for a single layer (dropout).
The results are presented in Table \ref{tab:ood_auc_stochastic_ibp}, and we find that even while using IBP as the bound-propagation method, our framework provides significantly stronger guarantees.

\begin{table}[t!]
  \centering
  \small
  \caption{
    Robust OOD Detection: MNIST vs EMNIST (MLP and LeNet) and CIFAR-10 vs CIFAR-100 (VGG-*).
    BP: Bound-Propagation (baseline), using IBP instead of \citet{bunel2020efficient} for bilinear operations; FL: Functional Lagrangian (ours).
    The reported times correspond to the median of the 500 samples.
    }
 \begin{tabular}{l cc c c | c c | c c| c}
  \toprule
        \multirow{2}{*}{OOD Task} &\multirow{2}{*}{Model} & \multirow{2}{*}{\#neurons} & \multirow{2}{*}{\#params} & \multirow{2}{*}{$\epsilon$} & \multicolumn{2}{c|}{Time (s)} & \multicolumn{2}{c|}{GAUC (\%)} & \multirow{2}{*}{AAUC (\%)} \\
        \cmidrule(lr){6-7} \cmidrule(lr){8-9}
         &  &  &  &  & BP & FL &BP &FL &\\
        \midrule
        \multirow{3}{*}{(E)MNIST} 
        & \multirow{3}{*}{MLP} & \multirow{3}{*}{256} & \multirow{3}{*}{2k} & 0.01 &1.1 & +13.1 & 55.4 & {\bf 67.5} & {\color{blue} 86.9} \\
         & &  &  & 0.03 &1.2 & +13.4 & 38.7 & {\bf 54.5} & {\color{blue} 88.6} \\
         & &  & & 0.05 &1.3 & +17.7 & 19.1 & {\bf 36.0} & {\color{blue} 88.8} \\ \midrule
        \multirow{3}{*}{(E)MNIST} 
        &\multirow{3}{*}{LeNet}  & \multirow{3}{*}{0.3M} & \multirow{3}{*}{0.1M} & 0.01 & 50.1& +13.1 & 0.0 & {\bf 28.4} & {\color{blue} 71.6} \\
         & & & & 0.03 & 54.7& +13.7 & 0.0 & {\bf 11.7} & {\color{blue} 57.6} \\
         & & & & 0.05 & 79.4& +24.8 & 0.0 & {\bf 2.3} & {\color{blue} 44.0} \\ \midrule
        \multirow{3}{*}{CIFAR}
        &VGG-16 &  3.0M & 83k & 0.001 & 426.4 & +21.4& 0.0 & {\bf 21.7} & {\color{blue} 60.9} \\
        &VGG-32 &  5.9M & 0.2M & 0.001 & 1035.2 & +21.3& 0.0 & {\bf 23.8} & {\color{blue} 64.7} \\
        &VGG-64 &  11.8M & 0.5M & 0.001 & 8549.7& +42.1 & 0.0 & {\bf 28.6} & {\color{blue} 67.4} \\
        \bottomrule
    \end{tabular}
    \label{tab:ood_auc_stochastic_ibp}
\end{table}

\subsection{Adversarial Robustness for Stochastic Neural Networks}
For the verification tasks considered in Section \ref{sec:experiments_adv_bnn}, we use 
IBP instead of the tighter LBP as the bound-propagation method and report 
results in Table \ref{tab:supp_bnn_adv}.
We find that, similar to Section \ref{sec:experiments_adv_bnn}, our framework is able to significantly improve on the guarantees the bound-propagation baseline is able to provide.
\begin{table}[t!]
  \centering
  \small
  \caption{
  Adversarial Robustness for different BNN architectures trained on MNIST from \cite{wicker2020probabilistic}. 
    BP: Bound-Propagation (baseline), using IBP instead of LBP for bilinear operations; FL: Functional Lagrangian (ours).
   The accuracy reported for FL and BP is the \% of samples we can certify as robust with  probability 1.
   For each model, we report results for the first 500 test-set samples. 
   }
   
  \begin{tabular}{c c c| c c  | c  c | c}
  \toprule
   \#layers  & $\epsilon$ &  \#neurons  & BP Acc. (\%) & FL Acc. (\%)  & BP Time (s) & FL Time (s) & Adv Acc (\%)\\
        \midrule
      &    & 128 & 43.8 & \textbf{65.2}  & 1.3 & +353.3 & {\color{blue} 82.6}\\
     1 & 0.025 & 256   & 40.6 & \textbf{64.6} & 1.4 & +431.3 & {\color{blue} 82.6} \\
      &  & 512 &   35.0 & \textbf{57.0} & 1.3 & +357.1 & {\color{blue}82.8} \\
     \hline 
     &  & 256 & 29.4 & \textbf{36.9} & 1.6 & +439.6 & {\color{blue} 79.4}\\
    2 &  0.001 & 512 & 46.0 & \textbf{63.4} & 1.7 & +433.8 & {\color{blue} 89.2}\\
     &  & 1024 & 18.4 & \textbf{19.6} & 1.6 & +440.9 & {\color{blue} 74.8}\\ 
    \bottomrule
    \end{tabular}
    \label{tab:bnn_adv}
\end{table}

\vfill

\end{document}